\documentclass[final]{cvpr}

\usepackage{times}
\usepackage{epsfig}
\usepackage{graphicx}
\usepackage{amsmath}
\usepackage{amssymb}
\usepackage{float}
\usepackage{booktabs}

\usepackage[pagebackref=true,breaklinks=true,colorlinks,bookmarks=false,citecolor=green]{hyperref}
\newcommand{\w}{{\rm\bf w}}     
\newcommand{\W}{\mathcal{W}}    
  
\newcommand{\x}{{\rm\bf x}}     
\newcommand{\X}{\mathcal{X}}

\newcommand{\Loss}{\mathcal{L}} 

\newcommand{\E}{\mathbb{E}}     
\newcommand{\z}{{\rm\bf z}}     
\newcommand{\Z}{\mathcal{Z}}    
\newcommand{\FINAL}[2][]{#2} 
\def\eg{\emph{e.g.}}
\def\ie{\emph{i.e.}}
\def\etal{\emph{et al.}}

\newcommand*{\affaddr}[1]{#1} 
\newcommand*{\affmark}[1][*]{\textsuperscript{#1}}

\def\confYear{CVPR 2021}

\begin{document}

\title{TediGAN: Text-Guided Diverse Face Image Generation and Manipulation}

\author{\hspace{-20pt}
Weihao Xia\affmark[1]
\qquad
Yujiu Yang\affmark[1]\thanks{Yujiu Yang is the corresponding author. This research was partially supported by the Key Program of National Natural Science Foundation of China under Grant No. U1903213 and the Guangdong Basic and Applied Basic Research Foundation (No. 2019A1515011387).}
\qquad
Jing-Hao Xue\affmark[2]
\qquad
Baoyuan Wu\affmark[3,4]\thanks{Baoyuan Wu is supported by the Natural Science Foundation of China under Grant No. 62076213, the University Development Fund of the Chinese University of Hong Kong, Shenzhen under Grant No. 01001810, and the Special Project Fund of Shenzhen Research Institute of Big Data under grant No. T00120210003.}
\\
{\tt\small \hspace{-10pt}
weihaox@outlook.com \hspace{0.5pt}
yang.yujiu@sz.tsinghua.edu.cn \hspace{0.5pt}
jinghao.xue@ucl.ac.uk \hspace{0.5pt}
wubaoyuan@cuhk.edu.cn}

\\
\affaddr{
\affmark[1]Tsinghua Shenzhen International Graduate School, Tsinghua University, China\\
\affmark[2]Department of Statistical Science, University College London, UK\\
\affmark[3]School of Data Science, Chinese University of Hongkong, Shenzhen, China\\
\affmark[4]Secure Computing Lab of Big Data, Shenzhen Research Institute of Big Data, Shenzhen, China}
}

\maketitle
\thispagestyle{empty}

\begin{abstract}
In this work, we propose TediGAN, a novel framework for multi-modal image generation and manipulation with textual descriptions.
The proposed method consists of three components: StyleGAN inversion module, visual-linguistic similarity learning, and instance-level optimization. 
The inversion module maps real images to the latent space of a well-trained StyleGAN.
The visual-linguistic similarity learns the text-image matching by mapping the image and text into a common embedding space.
The instance-level optimization is for identity preservation in manipulation.
Our model can produce diverse and high-quality images with an unprecedented resolution at $\text{1024}^2$.
Using a control mechanism based on style-mixing, our TediGAN inherently supports image synthesis with multi-modal inputs, such as sketches or semantic labels, with or without instance guidance.  
To facilitate text-guided multi-modal synthesis, we propose the Multi-Modal CelebA-HQ, a large-scale dataset consisting of real face images and corresponding semantic segmentation map, sketch, and textual descriptions.
Extensive experiments on the introduced dataset demonstrate the superior performance of our proposed method. 
Code and data are available at~\FINAL{\texttt{\small{https://github.com/weihaox/TediGAN}}}.
\end{abstract}

\section{Introduction}
\label{sec:intro}

\begin{figure}[th]
\centering
\includegraphics[width=1.0\linewidth]{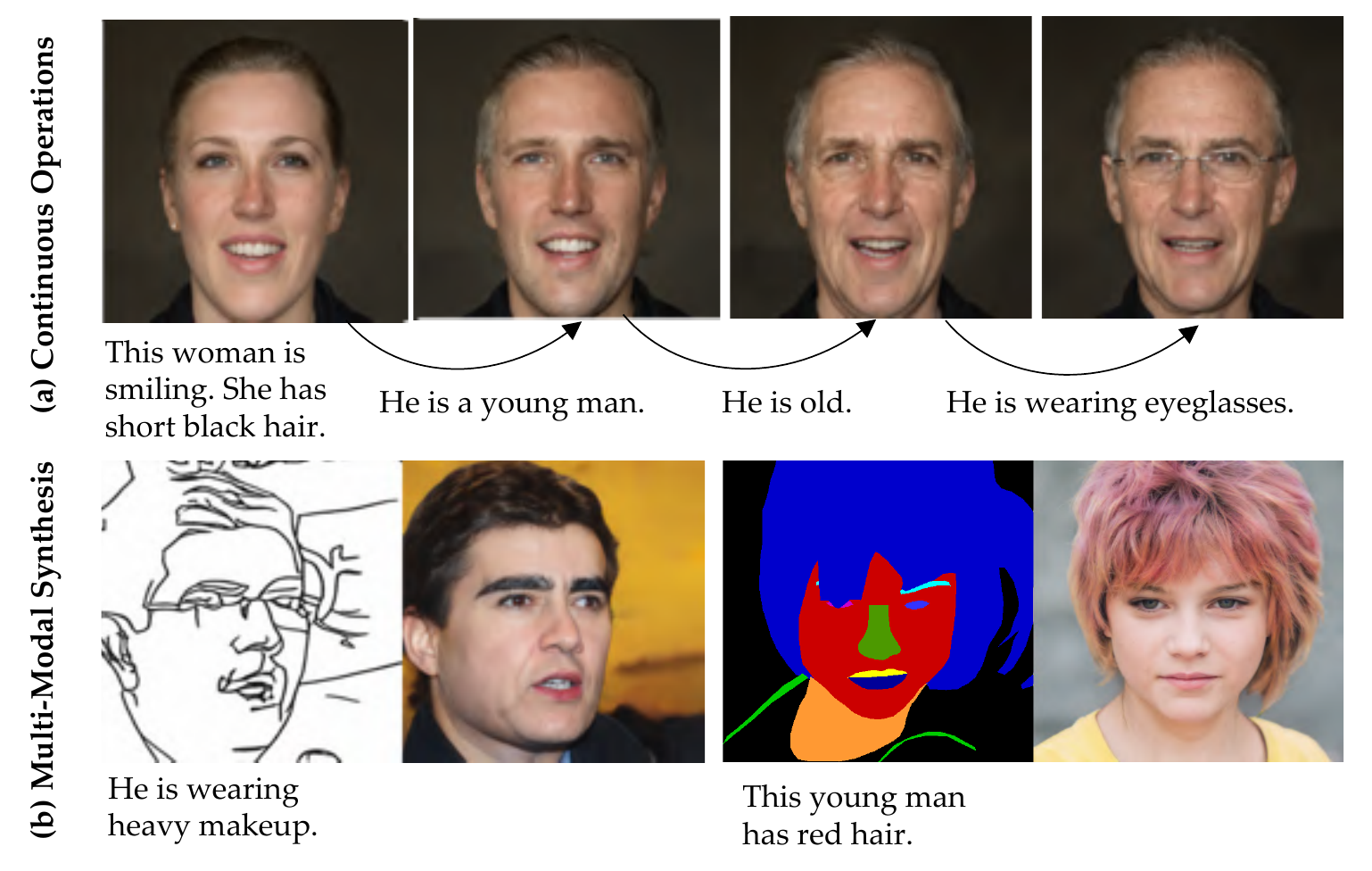}
\caption{Our TediGAN is the first method that unifies text-guided image generation and manipulation into one same framework, leading to naturally continuous operations from generation to manipulation (a), and inherently supports image synthesis with multi-modal inputs (b), such as sketches or semantic labels with or without instance (texts or real images) guidance.
}
\label{fig:teaser}
\end{figure}
How to create or edit an image of the desired content without tedious manual operations is a difficult but meaningful task.
To make image generation and manipulation more readily and user-friendly, recent studies have been focusing on the image synthesis conditioned on a variety of guidance, such as sketch~\cite{ghosh2019isketchnfill,xia2019sketch}, semantic label~\cite{isola2017image,wang2018high}, or textual description~\cite{nam2018tagan,xu2018attngan}.
Despite the success of its label and sketch counterparts, most state-of-the-art text-guided image generation and manipulation methods are only able to produce low-quality images~\cite{reed2016generative,dong2017semantic}. 
Those aiming at generating high-quality images from texts typically design a multi-stage architecture and train their models in a progressive manner. 
To be more specific, there are usually three stages in the main module, and each stage contains a generator and a discriminator. Three stages are trained at the same time, and progressively generate images of three different scales, \ie, \(\text{64}^2 \to \text{128}^2 \to \text{256}^2\).
The initial image with rough shape and color would be refined to a high-resolution one. 
However, the multi-stage training process is time-consuming and cumbersome, making the aforementioned methods unfeasible for higher resolution.
Furthermore, the pretrained text-image matching model they used fails to exploit attribute-level cross-modal information and leads to mismatched attributes when generating images from texts~\cite{xu2018attngan,li2019control,zhu2019dmgan,cheng2020rifegan}, or undesired changes of irrelevant attributes when manipulating images~\cite{dong2017semantic,nam2018tagan,li2020manigan,li2020lightweight}.

\begin{figure}[t]
\centering
\includegraphics[width=0.95\linewidth]{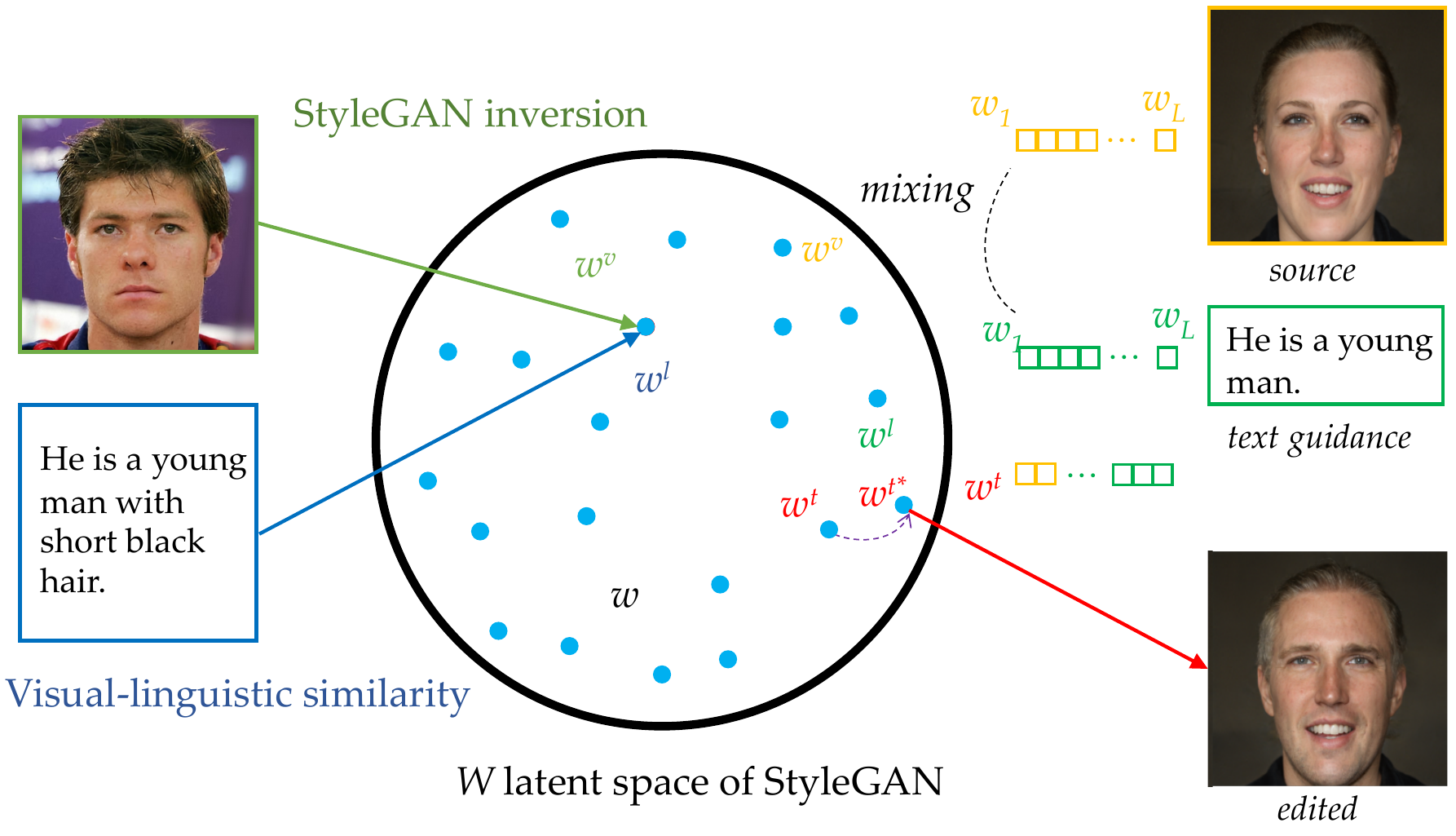}
\caption{{Projecting Multi-Modal Embedding into the $\mathcal{W}$ Space of StyleGAN.} 
Taking visual and linguistic embedding for example, the left illustrates visual-linguistic similarity learning, where the visual embedding $\w^v$ and linguistic embedding $\w^l$ are expected to be close enough. 
The right demonstrates text-guided image manipulation.
Given a source image and a text guidance, we first get their embedding $\w^v$ and $\w^l$ in $\mathcal{W}$ space through corresponding encoders. We then perform style mixing for target layers and get the target latent code $\w^t$. The final $\w^{t*}$ is obtained through instance-level optimization. The edited image can be generated from the StyleGAN generator.
}
\label{fig:mapping}
\end{figure}

Recent progress on generative adversarial networks (GANs) has established an entirely different image generation paradigm that achieves phenomenal quality, fidelity, and realism.
StyleGAN~\cite{karras2019style}, one of the most notable GAN frameworks, introduces a novel style-based generator architecture and can produce high-resolution images with unmatched photorealism.
Some recent work~\cite{karras2019style} has demonstrated that the intermediate latent space $\W$ of StyleGAN, inducted from a learned piece-wise continuous mapping, yields less entangled representations and offers more feasible manipulation.
The superior characteristics of $\W$ space appeal to numerous researchers to develop advanced GAN inversion techniques~\cite{xia2021survey,bau2019inverting,abdal2019image2stylegan} to invert real images back into the StyleGAN's latent space and perform meaningful manipulation.
The most popular way~\cite{zhu2020indomain,richardson2020encoding} is to train an additional encoder to map real images into the $\W$ space, which leads to not only faithful reconstruction but also semantically meaningful editing. 
Furthermore, it is easy to introduce the hierarchically semantic property of the $\W$ space to any GAN model by simply learning an extra mapping network before a fixed, pretrained StyleGAN generator.
We thoroughly investigated the existing GAN inversion methods, and found all is about how to map images into the latent space of a well-trained GAN model.
The other modalities like texts, however, have not received any attention.

In this paper, for the first time, we propose a GAN inversion technique that can map multi-modal information, \eg, texts, sketches, or labels, into a common latent space of a pretrained StyleGAN.
Based on that, we propose a very simple yet effective method for \textcolor{blue}{Te}xt-guided \textcolor{blue}{di}verse image generation and manipulation via \textcolor{blue}{GAN} (abbreviated \textcolor{blue}{TediGAN}). 
Our proposed method introduces three novel modules. The first StyleGAN inversion module learns the inversion where an image encoder can map a real image to the $\W$ space, 
while the second visual-linguistic similarity module learns linguistic representations that are consistent with the visual representations by projecting both into a common $\W$ space, as shown in Figure~\ref{fig:mapping}.
The third instance-level optimization module is to preserve the identity after editing, which can precisely manipulate the desired attributes consistent with the texts while faithfully reconstructing the unconcerned ones.
Our proposed method can generate diverse and high-quality results 
with a resolution up to $\text{1024}^2$,
and inherently support image synthesis with multi-modal inputs, such as sketches or semantic labels with or without instance (texts or real images) guidance.
Due to the utilization of a pretrained StyleGAN model, our method can provide the lowest effect guarantee, \ie, our method can always produce pleasing results no matter how uncommon the given text or image is.
Furthermore, to fill the gaps in the text-to-image synthesis dataset for faces, we create the Multi-Modal CelebA-HQ dataset to facilitate the research community.
Following the format of the two popular text-to-image synthesis datasets, \ie, CUB~\cite{wah2011caltech} for birds and COCO~\cite{lin2014microsoft} for natural scenes, we create ten unique descriptions for each image in the CelebA-HQ~\cite{karras2017progressive}. 
Besides real faces and textual descriptions, the introduced dataset also contains the label map and sketch for the text-guided generation with multi-modal inputs.

In summary, this work has the following contributions:
\begin{itemize} 
\item 
We propose a unified framework that can generate diverse images given the same input text, or text with image for manipulation, allowing the user to edit the appearance of different attributes interactively.

\item We propose a GAN inversion technique that can map multi-modal information into a common latent space of a pretrained StyleGAN where the instance-level image-text alignment can be learned.
\item We introduce the Multi-Modal CelebA-HQ dataset, consisting of multi-modal face images and corresponding textual descriptions, to facilitate the community.
\end{itemize}

\begin{figure*}[th]
\centering
\begin{minipage}[t]{0.33\textwidth}
\centering
\includegraphics[width=0.95\linewidth]{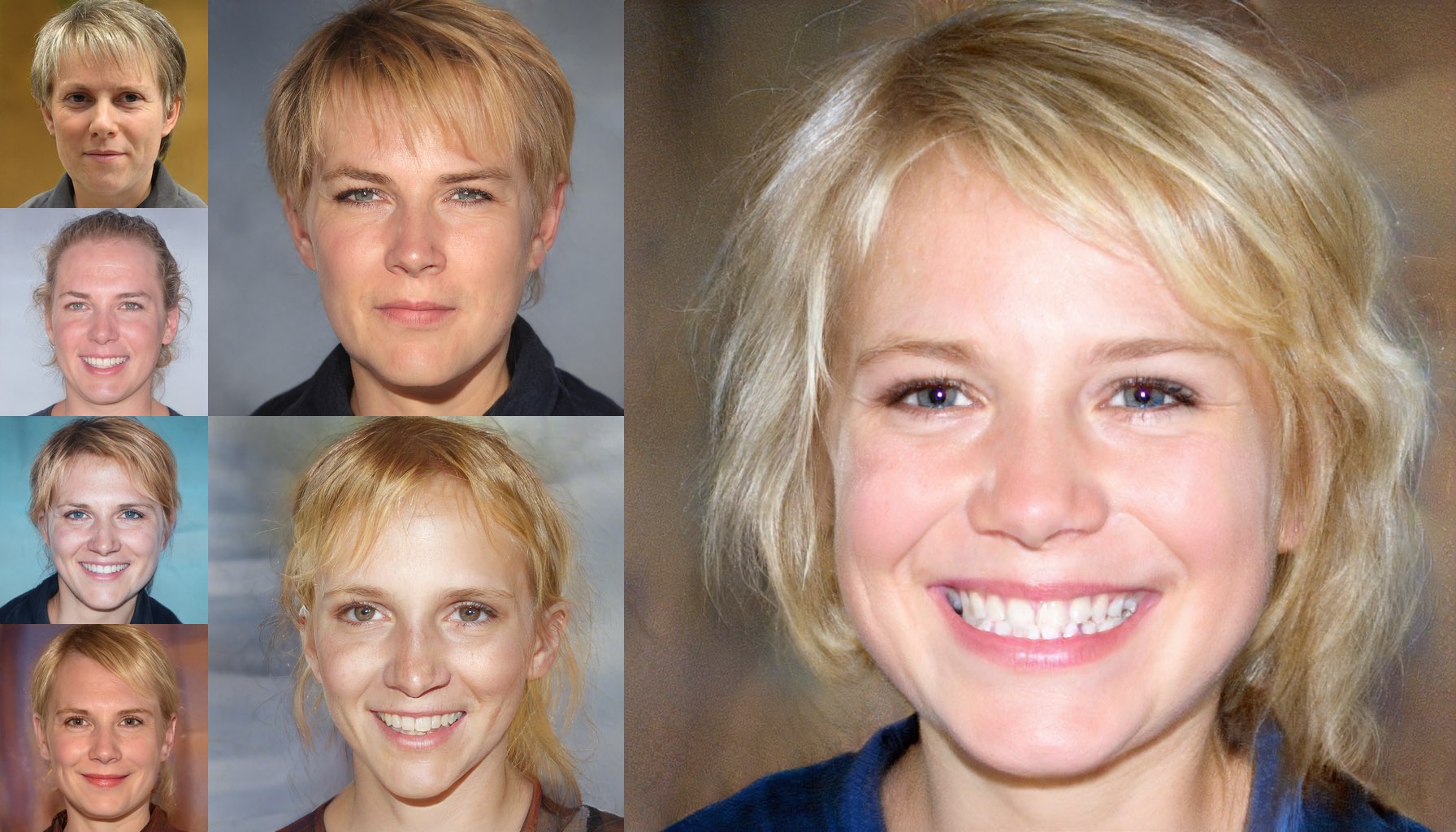}\\
\vspace{-5.0pt}\fontsize{7.0pt}{\baselineskip}\selectfont{\textit{a smiling young woman with short blonde hair}}
\end{minipage}
\begin{minipage}[t]{0.33\textwidth}
\centering
\includegraphics[width=0.95\linewidth]{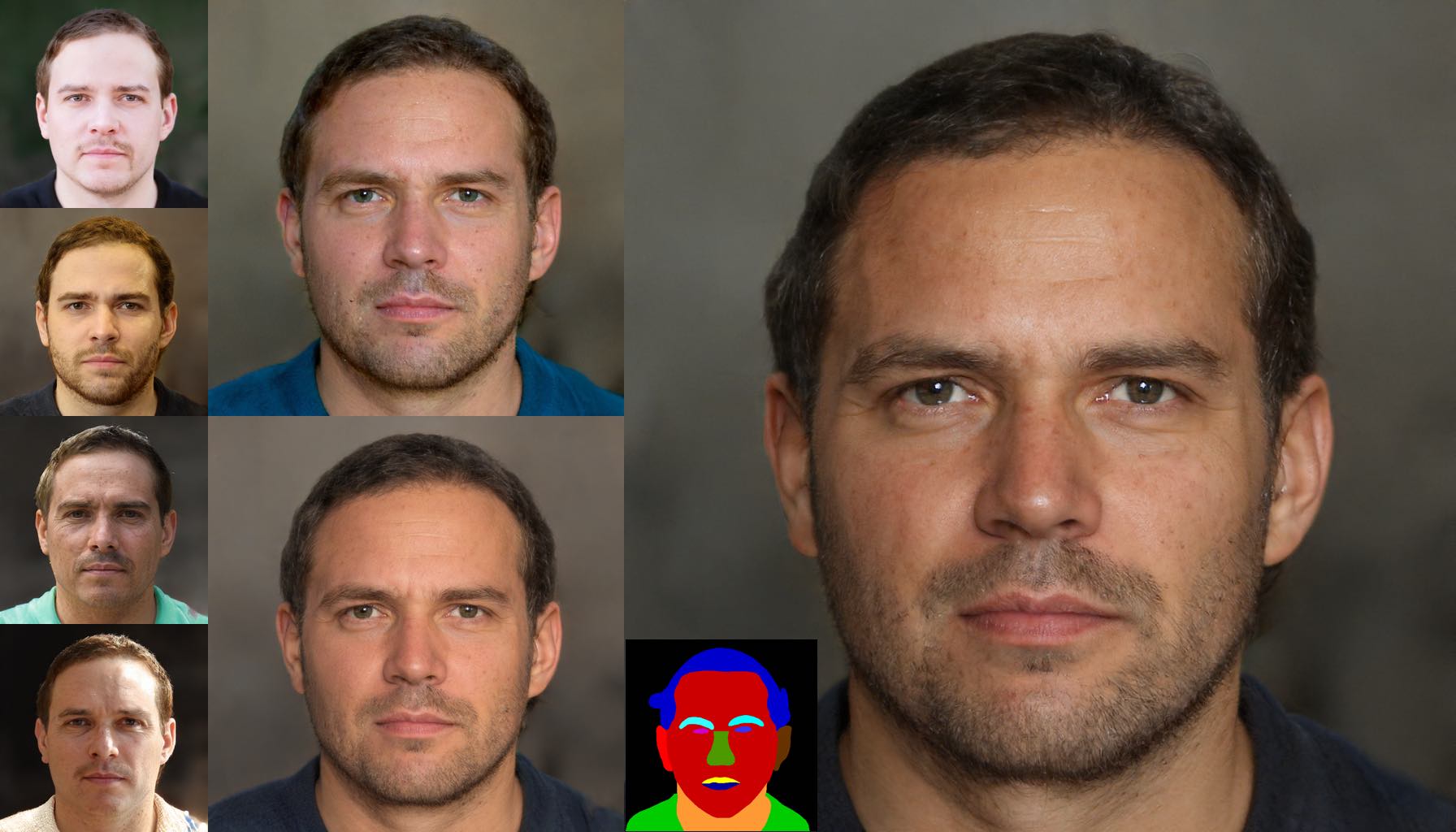}\\ 
\vspace{-5.0pt}\fontsize{7.0pt}{\baselineskip}\selectfont{\textit{he is young and wears beard}}
\end{minipage}
\begin{minipage}[t]{0.33\textwidth}
\centering
\includegraphics[width=0.95\linewidth]{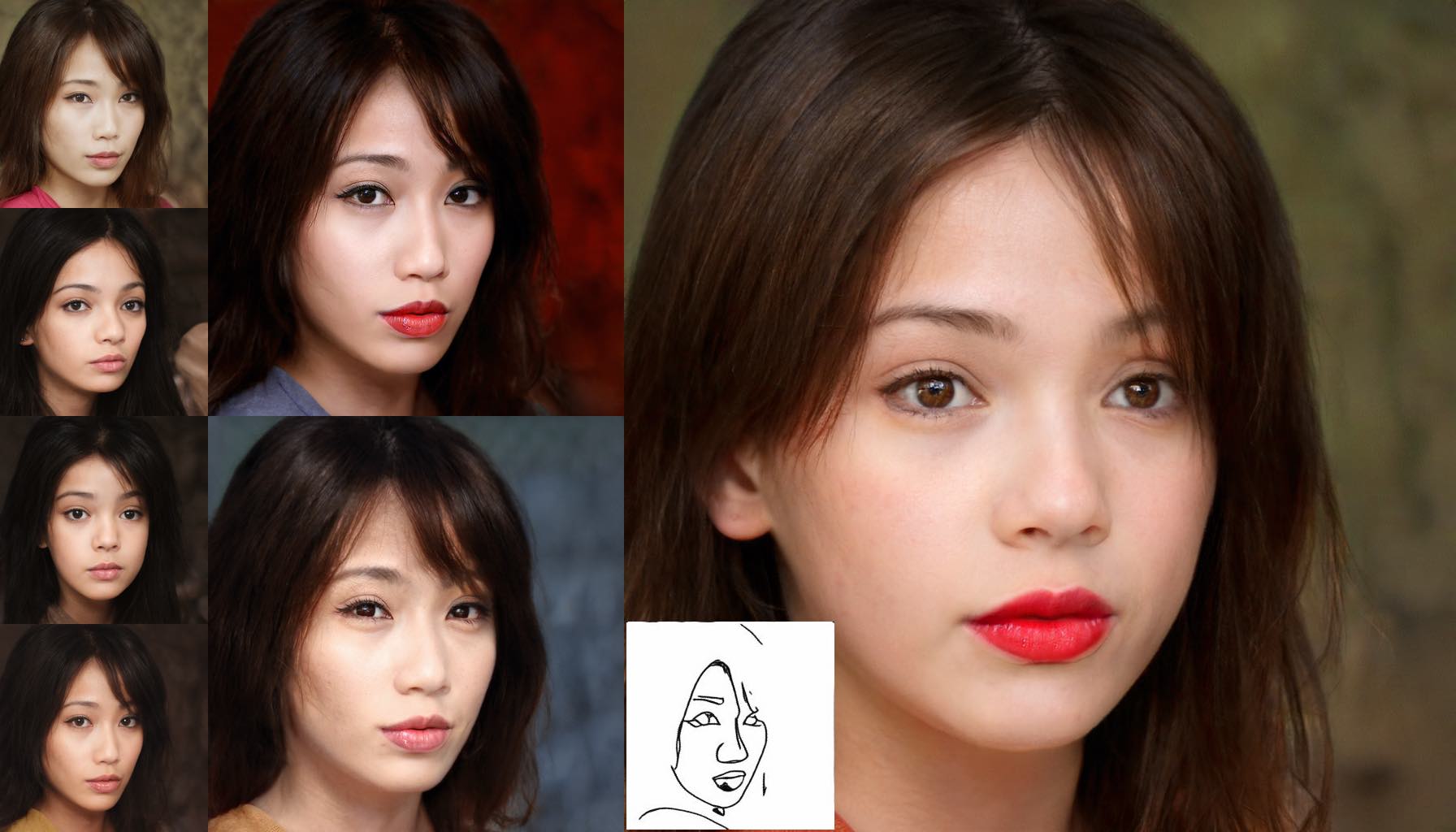}\\
\vspace{-5.0pt}\fontsize{7.0pt}{\baselineskip}\selectfont{\textit{a young woman with long black hair}}
\end{minipage}
\caption{Diverse High-Resolution Results from Text. 
Our method can achieve text-guided diverse image generation and manipulation up to an unprecedented resolution at $\text{1024}^2$.}
\label{fig:high-res-gene}
\end{figure*}

\section{Related Work}
\label{sec:related_work}

\paragraph{Text-to-Image Generation.}
There are basically two categories of GAN-based text-to-image generation methods.
The first category produces images from texts directly by one generator and one discriminator.
For example, Reed~\etal~\cite{reed2016generative} propose to use conditional GANs to generate plausible images from given text descriptions. 
Tao~\etal~\cite{tao2020dfgan} propose a simplified backbone that generates high-resolution images directly by Wasserstein distance and fuses the text information into visual feature maps to improve the image quality and text-image consistency.
Despite the plainness and conciseness, the one-stage models produce dissatisfied results in terms of both photo-realism and text-relevance in some cases.
Thus, another thread of research focuses on multi-stage processing.
Zhang~\etal~\cite{zhang2017stackgan} stack two GANs to generate high-resolution images from text descriptions through a sketch-refinement process.
They further propose a three-stage architecture~\cite{zhang2018stackgan++} that stacks multiple generators and discriminators, where multi-scale images are generated progressively in a course-to-fine manner.
Xu~\etal~\cite{xu2018attngan} improve the work of~\cite{zhang2018stackgan++} from two aspects. 
First, they introduce attention mechanisms to explore fine-grained text and image representations.
Second, they propose a Deep Attentional Multimodal Similarity Model (DAMSM) to compute the similarity between the generated image and the sentence.
The subsequent studies basically follow the framework of~\cite{xu2018attngan} and have proposed several variants by introducing different mechanisms like attention~\cite{li2019control} or memory writing gate~\cite{zhu2019dmgan}.
However, the multi-stage frameworks produce results that look like a simple combination of visual attributes from different image scales.

\vspace{-5pt}
\paragraph{Text-Guided Image Manipulation.}
Similar to text-to-image generation, manipulating given images using texts also produces results that contain desired visual attributes. 
Differently, the modified results should only change certain parts and preserve text-irrelevant contents of the original images. 
For example, Dong~\etal~\cite{dong2017semantic} propose an encoder-decoder architecture to modify an image according to a given text.
Nam~\etal~\cite{nam2018text} disentangle different visual attributes by introducing a text-adaptive discriminator, which can provide finer training feedback to the generator. 
Li \etal~\cite{li2020manigan} introduce a multi-stage network with a novel text-image combination module to produce high-quality results.
Similar to text-to-image generation, the text-based image manipulation methods with the best performance are basically based on the multi-stage framework.
Different from all existing methods, we propose a novel framework that unifies text-guided image generation and manipulation methods and can generate high-resolution and diverse images \texttt{directly} without multi-stage processing.

\vspace{-5pt}
\paragraph{Image-Text Matching.} 
One key of text-guided image generation or manipulation is to match visual attributes with corresponding words. To do this, current methods usually provide explicit word-level training feedback from the elaborately-designed discriminator~\cite{li2020manigan,li2020lightweight}.
There is also a rich line of work proposed to address a related direction named image-text matching, or visual-semantic alignment, aiming at exploiting the matching relationships and making the corresponding alignments between text and image.
Most of them can be categorized into two-branch deep architecture according to the granularity of representations for both modalities, \ie, global~\cite{kiros2014unifying,mao2014deep,ma2015multimodal} or local~\cite{karpathy2014deep,karpathy2015deep,lee2018stacked} representations.
The first category employs deep neural networks to extract the global features of both modalities, based on which their similarities are measured~\cite{mao2014deep}.
Another thread of work performs instance-level image-text matching~\cite{nam2017dual,lee2018stacked,song2019polysemous}, learning the correspondences between words and image regions~\cite{karpathy2015deep}.

\begin{figure*}[ht]
\centering
\includegraphics[width=1.0\linewidth]{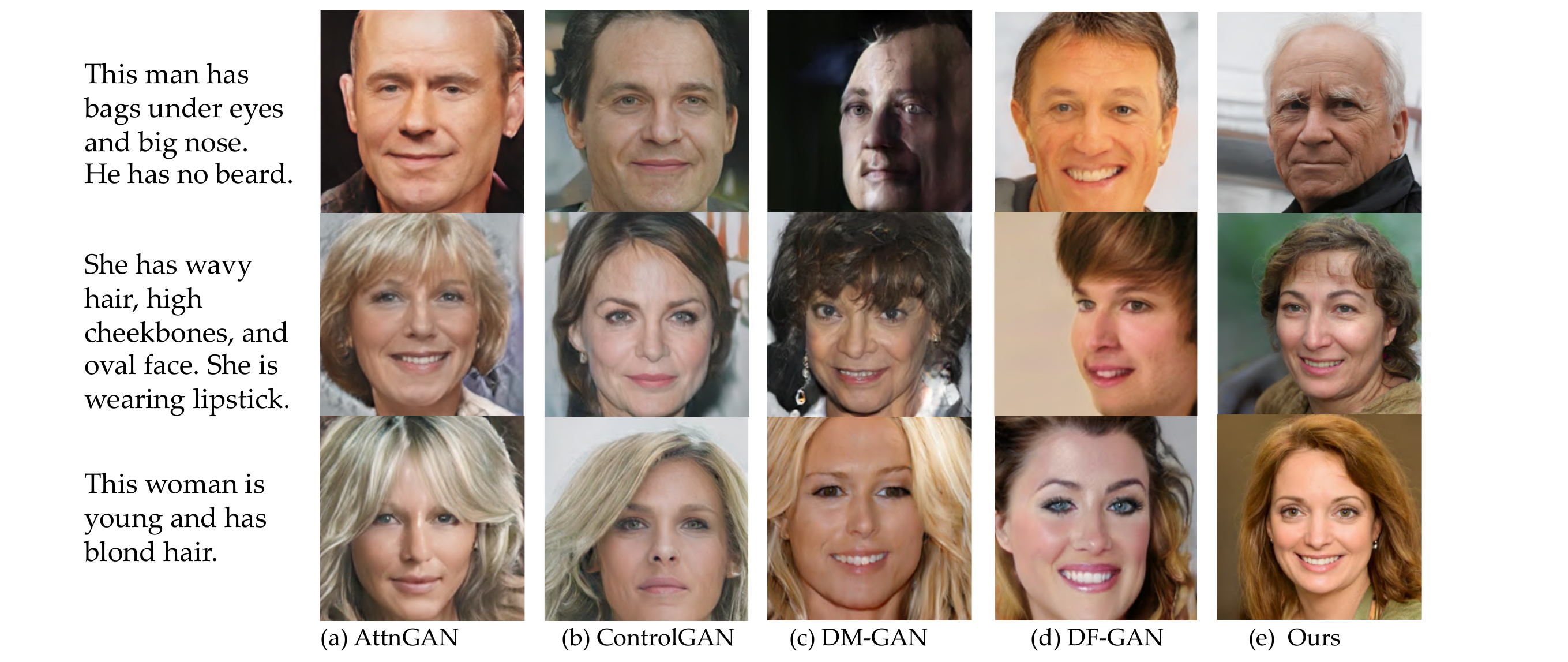}
\caption{Comparison of Text-to-Image Generation on Our Multi-modal CelebA-HQ dataset.}
\label{fig:comp_gen}
\end{figure*}

\section{The TediGAN Framework}
\label{sec:method}
We first learn the inversion, \ie, training an image encoder to map the real images to the latent space such that all codes produced by the encoder can be recovered at both the pixel-level and the semantic-level. 
We then use the hierarchical characteristic of $\W$ space to learn the text-image matching by mapping the image and text into the same joint embedding space.
To preserve identity in manipulation, we propose an instance-level optimization, involving the trained encoder as a regularization to better reconstruct the pixel values without affecting the semantic property of the inverted code. 

\subsection{StyleGAN Inversion Module}
\label{subsec:gan-inversion}
The inversion module aims at training an image encoder that can map a real face image to the latent space of a fixed StyleGAN model pretrained on the FFHQ dataset~\cite{karras2019style}.
The reason we invert a trained GAN model instead of training one from scratch is that, in this way, we can go beyond the limitations of a paired text-image dataset. 
The StyleGAN is trained in an unsupervised setting and covers much higher quality and wider diversity, which makes our method able to produce satisfactory edited results with images in the wild. 
In order to facilitate subsequent alignment with text attributes, our goal for inversion is not only to reconstruct the input image by pixel values but also to acquire the inverted code that is semantically meaningful and interpretable~\cite{shen2020interpreting,yang2019semantic}.

Before introducing our method, we first briefly establish problem settings and notations.
A GAN model typically consists of a generator $G(\cdot): \Z\rightarrow\X$ to synthesize fake images and a discriminator $D(\cdot)$ to distinguish real data from the synthesized. 
In contrast, GAN inversion studies the reverse mapping, which is to find the best latent code $\z^{*}$ by inverting a given image $\x$ to the latent space of a well-trained GAN.
A popular solution is to train an additional encoder $E_v(\cdot): \X\rightarrow\Z$~\cite{zhu2016generative,bau2019seeing} (subscript $v$ means visual).
To be specific, a collection of latent codes $\z^{s}$ are first randomly sampled from a prior distribution, \eg, normal distribution, and fed into $G(\cdot)$ to get the synthesis $\x^{s}$ as the training pairs.
The introduced encoder $E_v(\cdot)$ takes $\x^{s}$ and $\z^{s}$ as inputs and supervisions respectively and is trained with
\begin{align}
  \min_{\Theta_{E_v}}\Loss_{E_v} = ||\z^{s} - E_v(G(\z^{s}))||_2^2, 
  \label{eq:conventional-encoder}
\end{align}
where $||\cdot||_2$ denotes the $l_2$ distance and $\Theta_{E_v}$ represents the parameters of the encoder $E_v(\cdot)$.

Despite of its fast inference, the aforementioned procedure simply learns a deterministic model with no regard to whether the codes produced by the encoder align with the semantic knowledge learned by $G(\cdot)$.
The supervision by only reconstructing $\z^{s}$ is not powerful enough to train $E_v(\cdot)$, and $G(\cdot)$ is actually not fully used to guide the training of $E_v(\cdot)$, leading to the incapability of inverting real images.
To solve these problems, we use a totally different strategy to train an encoder for GAN inversion as in~\cite{zhu2020indomain}. 
There are two main differences compared with the conventional framework:
(a) the encoder is trained with real images rather than with synthesized images, making it more applicable to real applications;  
(b) the reconstruction is at the image space instead of latent space, which provides semantic knowledge and accurate supervision and allows integration of powerful image generation losses such as perceptual loss~\cite{johnson2016perceptual} and LPIPS~\cite{zhang2018unreasonable}.
Hence, the training process can be formulated as
\begin{small}
\begin{align}
  &\begin{aligned}
    \min_{\Theta_{E_v}}\Loss_{E_v} \!=\! ||\x - G(E_v(\x))||_2^2\ &\!+\!\lambda_{1} ||F(\x) \!-\! F(G(E_v(\x)))||_2^2\ \\
    &\!-\!\lambda_{2}\E[D_v(G(E_v(\x)))], \label{eq:encoder}
  \end{aligned} \\
  &\begin{aligned}
    \min_{\Theta_{D_v}}\Loss_{D_v} = \!\E[D_v(G(E_v(\x)))] \!- \!\E[D_v(\x)] \! + \!\frac{\lambda_3}{2} {\E}[||\nabla_{{\x}}D_v(\x)||_2^2],
  \end{aligned} \label{eq:discriminator}
\end{align}
\end{small}
where $\Theta_{E_v}$ and $\Theta_{D_v}$ are learnable parameters, 
$\lambda_{1}$, $\lambda_{2}$, and $\lambda_{3}$ are the hyper-parameters,
and $F(\cdot)$ denotes the VGG feature extraction model.

Through the learned image encoder, we can map a real image into the $\mathcal{W}$ space.
The obtained code is guaranteed to align with the semantic domain of the StyleGAN generator and can be further utilized to mine cross-modal similarity between the image-text instance pairs.

\begin{figure*}[ht]
\centering
\includegraphics[width=0.90\linewidth]{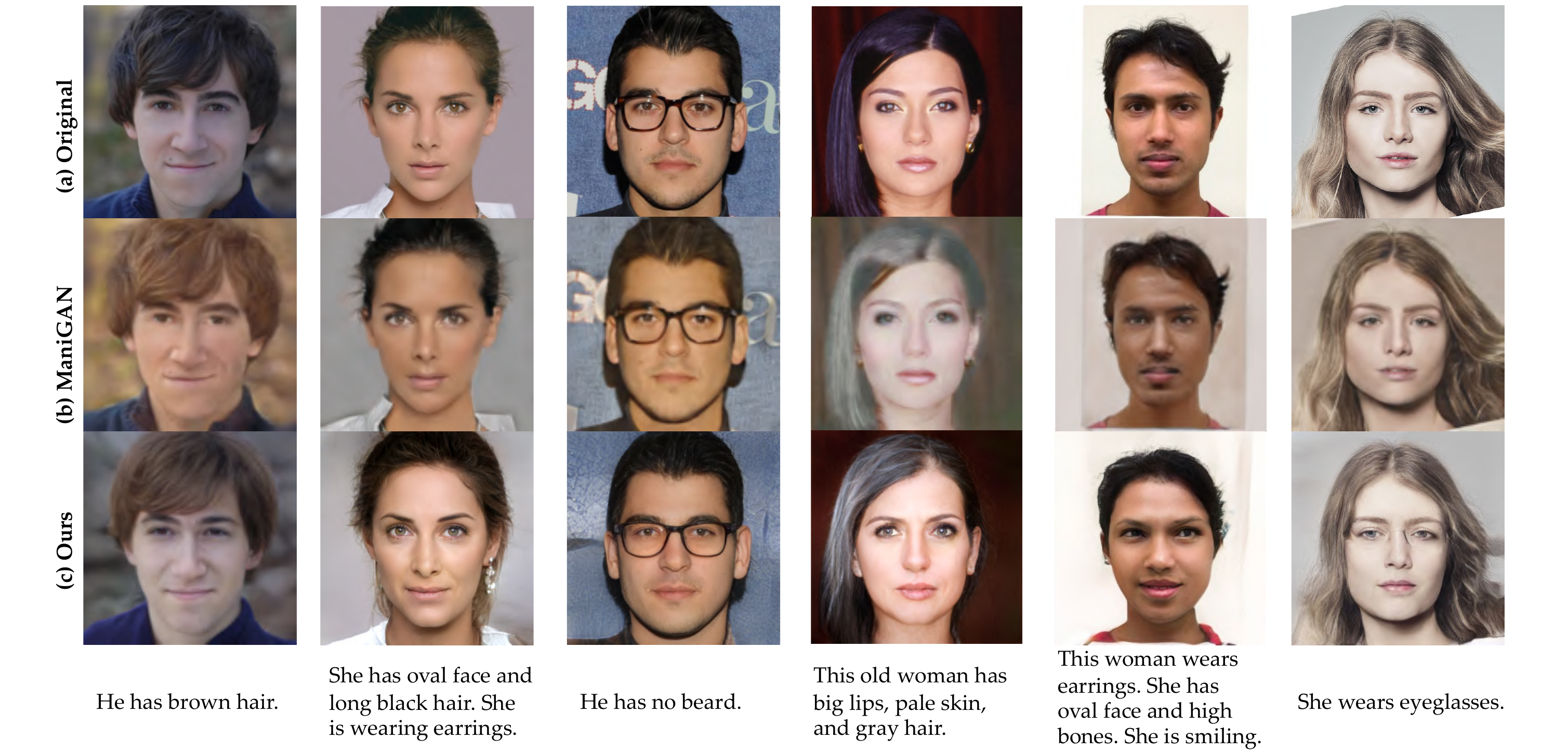}
\caption{Qualitative Comparison of Image Manipulation using Natural Language Descriptions.}
\label{fig:comp_man}
\end{figure*}

\subsection{Visual-Linguistic Similarity Learning}
\label{subsec:vls-model}

Once the inversion module is trained, given a real image, we can map it into the $\mathcal{W}$ space of StyleGAN. 
The next problem is how to train a text encoder that learns the associations and alignments between image and text.
Instead of training a text encoder in the same way as the image encoder or the aforementioned DAMSM, we propose a visual-linguistic similarity module to project the image and text into a common embedding space, \ie, the $\W$ space, as shown in Figure~\ref{fig:mapping}.
Given a real image and its descriptions, we encode them into the $\W$ space by using the previously trained image encoder and a text encoder. 
The obtained latent code is the concatenation of $L$ different $C$-dimensional $\w$ vectors, one for each input layer of StyleGAN.
The multi-modal alignment can be trained with 
\begin{align}
  \min_{\Theta_{E_l}}\Loss_{E_l} = ||\sum_{i=1}^{L} p_i (\w^v_i - \w^l_i)||_2^2, 
  \label{eq:reg}
\end{align}
where $\Theta_{E_l}$ represents the parameters of the text encoder $E_l(\cdot)$ and subscript $l$ means linguistic;
$\w^v, \w^l \in \W^{L \times C}$ are the obtained image embedding and text embedding; 
$\w^v = f(E_v(\x))$ is the projected code of the image embedding $\z$ in the input latent space $\Z$ using a non-linear mapping network $f:\Z \to \W$; $\w^l$ shares a similar definition;
$\w^v$ and $\w^l$ are with the same shape $L \times C$, meaning to have $L$ \textit{layers} and each with a $C$-\textit{dimensional} latent code;
and $p_i$ is the weight of $i$-th layer in the latent code.

Compared with DAMSM, our proposed module is lightweight and easy to train. More importantly, this module achieves instance-level alignment~\cite{Wang2020CVSE}, \ie, learning correspondences between visual and linguistic attributes, by leveraging the disentanglability of StyleGAN.
The text encoder is trained with the proposed visual-linguistic similarity loss together with the pairwise ranking loss~\cite{kiros2014unifying,dong2017semantic}, which is omitted from Equation~\ref{eq:reg}.

\subsection{Instance-Level Optimization}
\label{subsec:instance-level-optimization}
One of the main challenges of face manipulation is the identity preservation. Due to the limited representation capability, learning a perfect reverse mapping with an encoder alone is not easy. 
To preserve identity, some recent methods~\cite{ververas2020slidergan,richardson2020encoding} incorporate a dedicated face recognition loss~\cite{deng2019arcface} to measure the cosine similarity between the output image and its source.
Different from their methods, for text-guided image manipulation, we implement an instance-level optimization module to precisely manipulate the desired attributes consistent with the descriptions while faithfully reconstructing the unconcerned ones.  
We use the inverted latent code $\z$ as the initialization, and the image encoder is included as a regularization to preserve the latent code within the semantic domain of the generator. 
To summarize, the objective function for optimization is
\begin{align}
  \begin{aligned}
    \z^{*} = \arg\min_{\z}\ ||\x - G(\z)||_2^2\ &+ \lambda_{1}^{\prime}||F(\x) - F(G(\z))||_2^2\ \\
    &+ \lambda_{2}^{\prime}||\z - E_v(G(\z))||_2^2,
  \end{aligned} \label{eq:optimization}
\end{align}
where $\x$ is the original image to manipulate, $\lambda_{1}^{\prime}$ and $\lambda_{2}^{\prime}$ are the loss weights corresponding to the perceptual loss and the encoder regularization term, respectively.

\begin{figure}[t]
\centering
\includegraphics[width=1.0\linewidth]{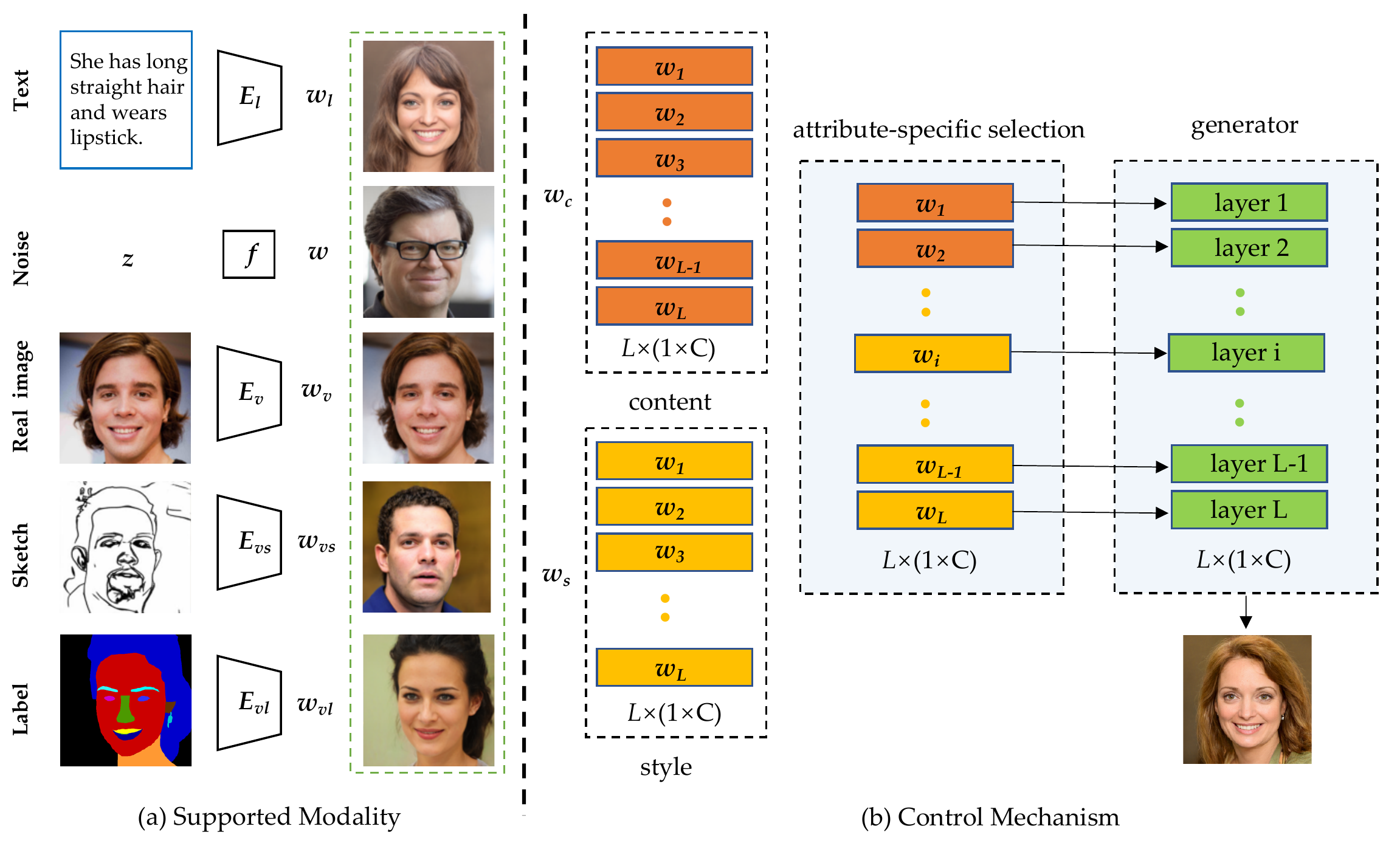}
\caption{Control Mechanism of Our TediGAN Framework.
Different layer in the StyleGAN generator represents different attributes. 
Changing the value of a certain layer would change the corresponding attributes of the image.
Since the texts and images are mapped into the common latent space, we can synthesize images with certain attributes by selecting attribute-specific layers.
The control mechanism mixes layers of the style code $\w^s$ by partially replacing corresponding layers of the content $\w^c$.
When $\w^s$ is a randomly sampled latent code, it is the text-to-image generation and when $\w^s$ is the image embedding, it performs text-guided image manipulation. 
}
\label{fig:control_mechanism}
\end{figure}

\subsection{Control Mechanism}
\label{subsec:control_mechanism}
\paragraph{Attribute-Specific Selection.} 
The two different tasks, \ie, text-to-image generation and text-guide image manipulation, are unified into one framework by our proposed control mechanism. 
Our mechanism is based on the style mixing of StyleGAN.
The layer-wise representation of StyleGAN learns disentanglement of semantic fragments (attributes or objects).
In general, different layer $\w_i$ represents different attributes and is fed into the $i$-th layer of the generator. 
Changing the value of a certain layer would change the corresponding attributes of the image.
As shown in Figure~\ref{fig:mapping}, given two codes with the same size $\w^c, \w^s \in \W^{\;L \times C}$ denoting content code and style code, this control mechanism selects attribute-specific layers and mixes those layers of $\w^s$ by partially replacing corresponding layers of $\w^c$. 
For text-to-image generation, the produced images should be consistent with the textual description, thus $\w^c$ should be the linguistic code, and randomly sampled latent code with the same size acts as $\w^s$ to provide diversity (results are shown in Figure~\ref{fig:diverse_image}). 
For text-guided image manipulation, $\w^c$ is the visual embedding while $\w^s$ is the linguistic embedding, the layers for mixing should be relevant to the text, for the purpose of modifying the relevant attributes only and keeping the unrelated ones unchanged.

\vspace{-5pt}
\paragraph{Supported Modality.} 
The style code $\w^s$ and content code $\w^c$ could be sketch, label, image, and noise, as shown in Figure~\ref{fig:control_mechanism}, which makes our TediGAN feasible for multi-modal image synthesis.
The control mechanism provides high accessibility, diversity, controllability, and accurateness for image generation and manipulation.
Due to the control mechanism, as shown in Figure~\ref{fig:teaser}, our method inherently supports continuous operations and multi-modal synthesis for sketches and semantic labels with descriptions.
To produce the diverse results, all we need to do is to keep the layers related to the text unchanged and replace the others with the randomly sampled latent code.
If we want to generate images from other modality with text guidance, take the sketch as an example, we can train an additional sketch image encoder $E_{vs}$ in the same way as training the real image encoder and leave the other parts unchanged.

\vspace{-5pt}
\paragraph{Layerwise Analysis.} 
The pre-trained StyleGAN we used in most experiments is to generate images of 256 $\times$ 256 (\ie, size 256), whose has 14 layers of the intermediate vector.
For a synthesis network trained to generate images of 512 $\times$ 512, the intermediate vector would be of shape (16, 512) (and (18, 512) for 1024 $\times$ 1024), where the number of the layers $L$ is determined by $2\log_2\mathrm{R}-2$ and $\mathrm{R}$ is the image size.
In general, layers in the generator at lower resolutions (\eg, 4 $\times$ 4 and 8 $\times$ 8) control high-level styles such as eyeglasses and head pose,
layers in the middle (\eg, as 16 $\times$ 16 and 32 $\times$ 32) control hairstyle and facial expression, while the final layers (\eg, 64 $\times$ 64 to 1024 $\times$ 1024) control color schemes and fine-grained details.
Based on empirical observations, we list the attributes represented by different layers of a 14-layer StyleGAN in Table~\ref{tab:layerwise_analysis}.
The layers from 11-14 represent micro features or fine structures, such as stubble, freckles, or skin pores, which can be regarded as the stochastic variation.
High-resolution images contain lots of facial details and cannot be obtained by simply upsampling from the lower-resolutions, making the stochastic variations especially important as they improve the visual perception without affecting the main structures and attributes of the synthesized image.

\begin{table}[th]
\centering
\caption{The Empirical Layerwise Analysis of a 14-layer StyleGAN Generator. The 13-th and 14-th layers are omitted since there is basically no visible difference.}
\scalebox{0.85}{
\begin{tabular}{cc|cc}
\toprule
$n$-th & attribute & $n$-th & attribute \\
\midrule
1 &eye glasses & 7 & hair color  \\
2 &head pose & 8 & face color  \\
3 &face shape & 9 & age \\
4 &hair length, nose, lip & 10 & gender \\
5 &cheekbones  & 11 & micro features \\
6 &chin & 12 & micro features \\
\bottomrule
\end{tabular}
}
\label{tab:layerwise_analysis}
\end{table}

\section{Experiments}
\label{sec:experiments}

\subsection{Experiments Setup}
\label{subsec:setup}

\begin{figure}[th]
\includegraphics[width=0.9\linewidth]{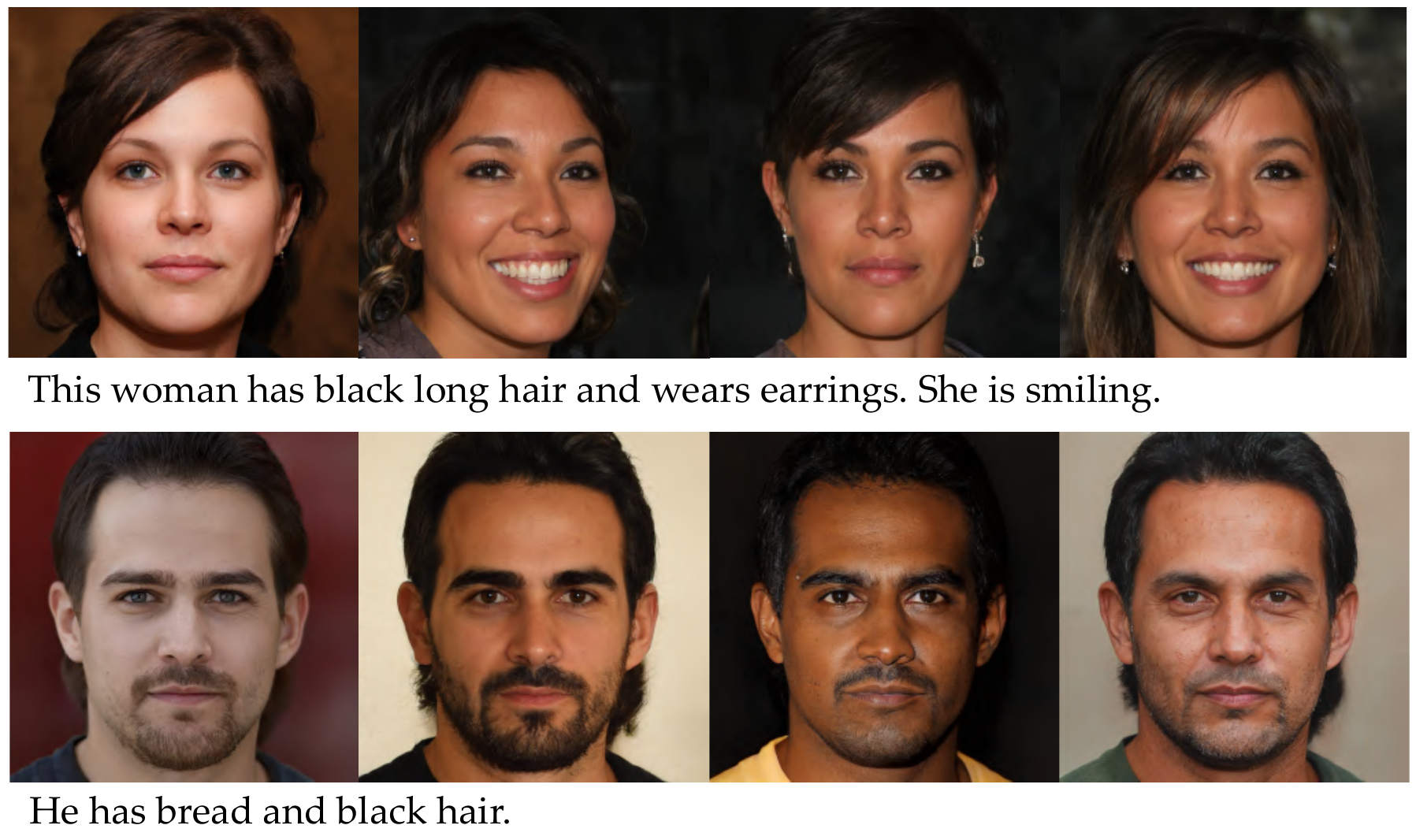}
\centering
\caption{Diverse Text-to-image Generation.}
\label{fig:diverse_image}
\end{figure}

\begin{table}[th]
\centering
\caption{{Quantitative Comparison of Text-to-Image Generation.} We use FID, LPIPS, accuracy (Acc.), and realism (Real.) to compare the state of the art and our method on the proposed Multi-modal CelebA-HQ dataset. $\downarrow$ means the lower the better while $\uparrow$ means the opposite.}
\scalebox{0.8}{
\begin{tabular}{c|ccccc}
\toprule
Method &FID $\downarrow$ &LPIPS $\downarrow$ & Acc. (\%) $\uparrow$ &Real. (\%) $\uparrow$\\
\midrule
AttnGAN~\cite{xu2018attngan} &125.98 &0.512 &14.2 &20.3 \\
ControlGAN~\cite{li2019control} &116.32 &0.522 &18.2 &22.5 \\
DFGAN~\cite{tao2020dfgan} &137.60 &0.581 &22.8 &25.5 \\
DM-GAN~\cite{zhu2019dmgan} &131.05 &0.544 &19.5 &12.8 \\
TediGAN &\textbf{106.37} &\textbf{0.456} &\textbf{25.3} &\textbf{31.7} \\
\bottomrule
\end{tabular}
}
\label{tab:quan_gen}
\end{table}

\paragraph{Datasets and Baseline Models.}
To achieve text-guided image generation and manipulation, the first step is to build a dataset that contains photo-realistic facial images and corresponding descriptions. 
We introduce the Multi-Modal CelebA-HQ dataset, a large-scale face image dataset that has 30,000 high-resolution face images, each having a high-quality segmentation mask, sketch, and descriptive text. 
We evaluate our proposed method on text and image partitions, comparing with state-of-the-art approaches AttnGAN~\cite{xu2018attngan}, ControlGAN~\cite{li2019control}, DM-GAN~\cite{zhu2019dmgan}, and DFGAN~\cite{tao2020dfgan} for image generation, and comparing with ManiGAN~\cite{li2020manigan} for image manipulation using natural language descriptions. 
All methods are retrained with the default settings on the proposed Multi-Modal CelebA-HQ dataset.

\vspace{-5pt}
\paragraph{Evaluation Metric.} 
For evaluation, there are four important aspects: image quality, image diversity, accuracy, and realism~\cite{li2019control,li2020lightweight}.
The quality of generated or manipulated images is evaluated through Fr\'echet Inception Distance (FID)~\cite{heusel2017gans}.
The diversity is measured by the Learned Perceptual Image Patch Similarity (LPIPS)~\cite{zhang2018unreasonable}. 
For image generation, the accuracy is evaluated by the similarity between the text and the corresponding generated image.
For manipulation, the accuracy is evaluated by whether the modified visual attributes of the synthetic image are aligned with the given description and text-irrelevant contents are preserved.
The accuracy and realism are evaluated through a user study, where the users are asked to judge which one is more photo-realistic, and more coherent with the given texts.
We test accuracy and realism by randomly sampling 50 images with the same conditions and collect more than 20 surveys from different people with various backgrounds.

\begin{table}[th]
\centering
\caption{{Quantitative Comparison of Text-Guided Image Manipulation.} We use FID, accuracy (Acc.), and realism (Real.) to compare with the state of the art ManiGAN~\cite{li2020manigan}.}
\scalebox{0.8}{
\begin{tabular}{c|cc|cc}
\toprule
&\multicolumn{2}{c}{CelebA} &\multicolumn{2}{c}{Non-CelebA} \\
Method 
&ManiGAN~\cite{li2020manigan} &Ours 
&ManiGAN~\cite{li2020manigan} &Ours \\
\midrule
FID $\downarrow$ &117.89 &\textbf{107.25} &143.39 &\textbf{135.47}  \\
Acc. (\%) $\uparrow$ &40.9 &\textbf{59.1} &12.8 &\textbf{87.2}  \\
Real. (\%) $\uparrow$ &36.2 &\textbf{63.8} &21.7 &\textbf{78.3}  \\
\bottomrule
\end{tabular}
}
\label{tab:quan_man}
\end{table}

\subsection{Comparison with State-of-the-Art Methods}
\label{subsec:comparison}

\subsubsection{Text-to-Image Generation}
\label{subsec:exp_gen}

\paragraph{Quantitative Comparison.}
In our experiments, we evaluate the FID and LPIPS on a large number of samples generated from randomly selected text descriptions. 
To evaluate accuracy and realism, we generate images from 50 randomly sampled texts using different methods. In a user study, users are asked to judge which one is the most photo-realistic and most coherent with the given texts. 
The results are demonstrated in Table~\ref{tab:quan_gen}. 
Compared with the state-of-the-arts, our method achieves better FID, LPIPS, accuracy, and realism values, which proves that our methods can generate images with the highest quality, diversity, photorealism, and text-relevance. 

\vspace{-5pt}
\paragraph{Qualitative Comparison.} 
Most existing text-to-image generation methods, as shown in Figure~\ref{fig:comp_gen}, can generate photo-realistic and text-relevant results.
However, some attributes contained in the text do not appear in the generated image, and the generated image looks like featureless paint and lacks details.
This ``featureless painterly'' look~\cite{karras2019style} would be significantly obvious and irredeemable when generating higher resolution images using the multi-stage training methods~\cite{xu2018attngan,li2019control,zhu2019dmgan}.
Furthermore, most existing solutions have limited diversity of the outputs, even if the provided conditions contain different meanings. For example, ``\textit{has a beard}" might mean a goatee, short or long beard, and could have different colors.
Our method can not only generate results with diversity but also realise the expectation to change where you want by using the control mechanism.
To produce diverse results, with the layers related to the text unchanged, the other layers could be replaced by any values sampled from the prior distribution. 
For example, as shown in the first row of Figure~\ref{fig:diverse_image}, the key visual attributes (\textit{women, black long hair, earrings, and smiling}) are preserved, while the other attributes, like haircuts, makeups, face shapes, and head poses, show a great degree of diversity.
The images in the second row illustrate more precise control ability. We keep the layers representing face shape and head pose the same and change the others. 
Figure~\ref{fig:high-res-gene} shows high-quality and diverse results with resolution at 1024 $\times$ 1024. 

\vspace{-5pt}
\subsubsection{Text-Guided Image Manipulation}
\label{subsec:exp_man}

\vspace{-5pt}
\paragraph{Quantitative Comparison.}
In our experiments, we evaluate the FID and conduct a user study on randomly selected images from both CelebA and Non-CelebA datasets with randomly chosen descriptions. 
The results are shown in Table~\ref{tab:quan_man}. Compared with ManiGAN~\cite{li2020manigan}, our method achieves better FID, accuracy, and realism. 
This indicates that our method can produce high-quality synthetic images, and the modifications are highly aligned with the given descriptions, while preserving other text-irrelevant contents. 

\vspace{-5pt}
\paragraph{Qualitative Comparison.} 
Figure~\ref{fig:comp_man} shows the visual comparisons between the recent method ManiGAN~\cite{li2020manigan} and ours.
As shown, the second row is to add earrings and change the face shape and hair style of the woman, our method completes this difficult case while ManiGAN fails to produce required attributes.
ManiGAN produces less satisfactory modified results: in some cases, the text-relevant regions are not modified and the text-irrelevant ones are changed.
Furthermore, since the StyleGAN we used is pretrained on a very large face dataset~\cite{karras2019style}, the latent space almost covers the full space of facial attributes, which makes our method robust for real images in the wild. 
The images in last two columns are results of out-of-distribution (Non-CelebA), \ie, images from other face dataset such as~\cite{chelnokova2014rewards, courset2018caucasian, yi2019apdrawinggan}, which illustrate that our method is prepared to produce pleasing results with images in the wild.

\section{Ablation Study and Discussion}
\label{sec:ablation}

\begin{figure}[t]
\includegraphics[width=0.95\linewidth]{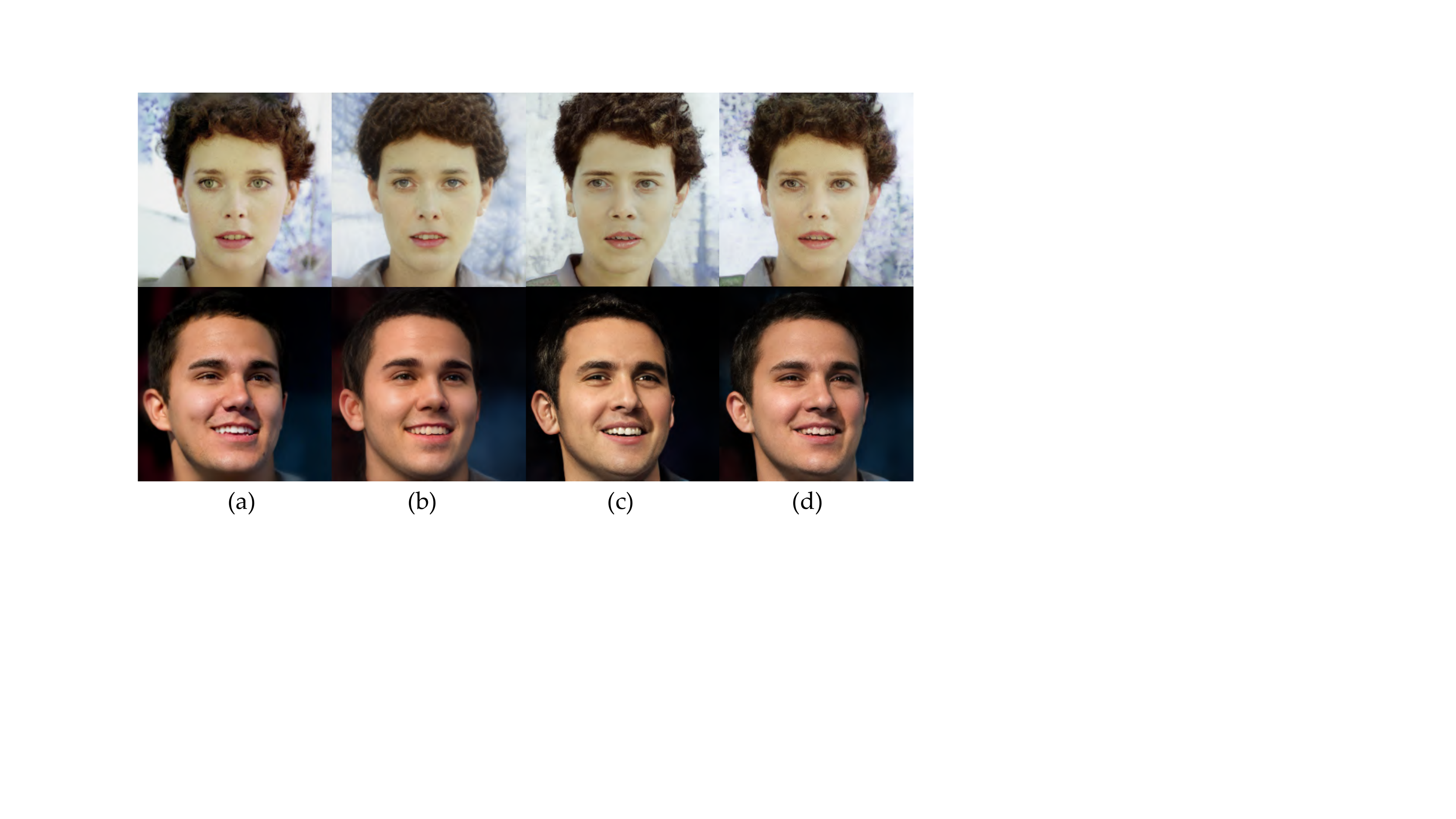}
\centering
\caption{{Inversion Results.} 
(a) original image; 
(b) inversion result of pSp~\cite{richardson2020encoding}; 
(c) inversion result of our image encoder (Section~\ref{subsec:gan-inversion}); 
(d) inversion results after optimization (Section~\ref{subsec:instance-level-optimization}).
}
\label{fig:inversion_result}
\end{figure}

\paragraph{Instance-Level Optimization.}
The comparison of with or without instance-level optimization is shown in Figure~\ref{fig:inversion_result}. 
As shown, the inversion results of the image encoder preserve all attributes of the original images, which is sufficient for text-to-image generation since there is no identity to preserve (Figure~\ref{fig:inversion_result} (c)). 
Manipulating a given image according to a text, however, should not change the unrelated attributes especially one's identity, which is preserved after the instance-level optimization (Figure~\ref{fig:inversion_result} (d)).  
We also compare with 
a recent inversion-based image synthesis method pSp~\cite{richardson2020encoding} that incorporates a dedicated recognition loss~\cite{deng2019arcface} during training.
Despite both preserving the identity, the optional instance-level optimization provides a non-deterministic way to refine the final results accordingly.

\begin{figure}[t]
\begin{center}
\includegraphics[width=0.95\linewidth]{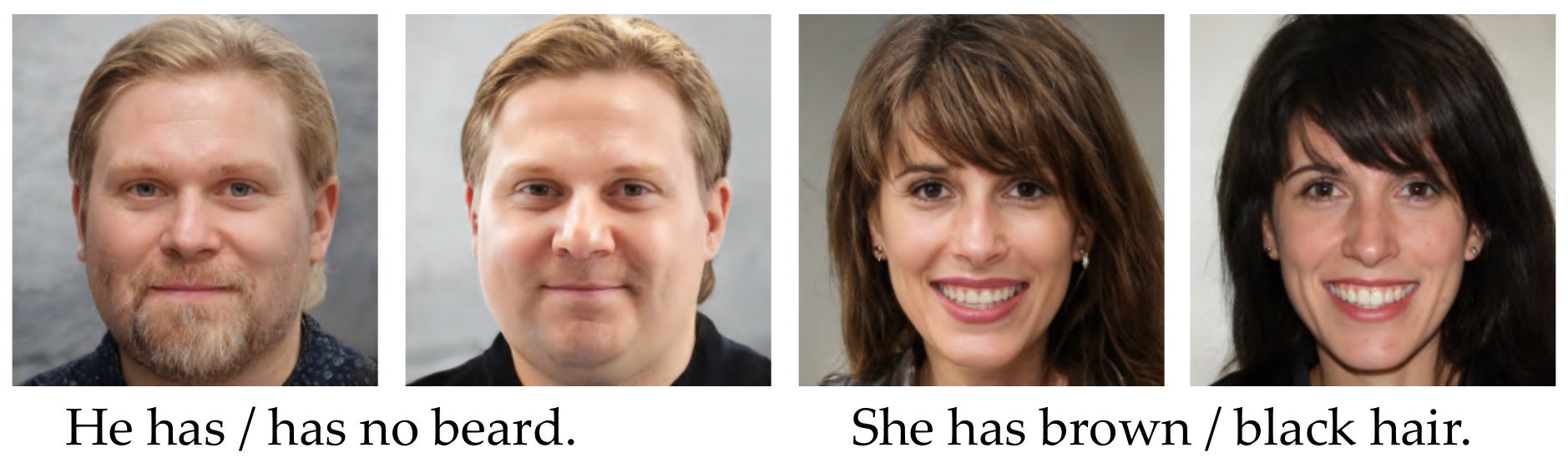}
\end{center}
\caption{Illustration of Near-miss Cases.}
\label{fig:near-miss}
\end{figure}

\vspace{-5pt}
\paragraph{Visual-Linguistic Similarity.}
The text encoder is trained using our visual-linguistic similarity and a very simple pairwise ranking loss~\cite{kiros2014unifying,dong2017semantic} to align text and image embedding.
Although the learned text embedding can handle near-miss cases, as shown in Figure~\ref{fig:near-miss}, we found this plain strategy sometimes may lead to insufficient disentanglement of attributes and mismatching of image-text alignment, leaving some room for improvement.

\vspace{-5pt}
\paragraph{Potential Issue with StyleGAN.}
In our experiments, we found that some unrelated attributes are unwantedly changed when we manipulate a given image according to a text description.
We thought it might be the problem of visual-linguistic similarity learning in the first place. 
However, when performing layer-wise style mixing on the inverted codes of two real images, the interference still occurs.
This means some facial attributes remain entangled in the $\mathcal{W}$ space, where different attributes should be orthogonal (meaning without affecting other attributes).
Another inherent defect of StyleGAN is that some attributes, such as hats, necklaces and earrings, are not well represented in its latent space. This makes our method perform less satisfactorily sometimes when adding or removing jewelry or accessories through natural language descriptions.

\section{Conclusion}
\label{sec:conclusion}
We have proposed a novel method for image synthesis using textual descriptions, which unifies two different tasks (text-guided image generation and manipulation) into the same framework and achieves high accessibility, diversity, controllability, and accurateness for facial image generation and manipulation.
Through the proposed multi-modal GAN inversion and large-scale multi-modal dataset, our method can effectively synthesize images with unprecedented quality.
Extensive experimental results demonstrate the superiority of our method, in terms of the effectiveness of image synthesis, the capability of generating high-quality results, and the extendability for multi-modal inputs.

{\small
\bibliographystyle{ieee}
\bibliography{reference}
}

\end{document}